\documentclass[journal]{IEEEtran}
\ifCLASSINFOpdf
   \usepackage[pdftex]{graphicx}
\else
   \usepackage[dvips]{graphicx}
\fi
\usepackage{amsmath}
\usepackage{amssymb}
\usepackage{bbold}
\usepackage{mathtools}
\usepackage{comment}
\usepackage{algorithm}
\usepackage{algpseudocode}
\usepackage{listings}

\usepackage{array}

\ifCLASSOPTIONcompsoc
  \usepackage[caption=false,font=normalsize,labelfont=sf,textfont=sf]{subfig}
\else
  \usepackage[caption=false,font=footnotesize]{subfig}
\begin{document}
%
\title{From Principal Subspaces to Principal Components with Linear Autoencoders}
%
%
%

\author{Elad~Plaut
}
\maketitle

\begin{abstract}
The autoencoder is an effective unsupervised learning model which is widely used in deep learning. It is well known that an autoencoder with a single fully-connected hidden layer, a linear activation function and a squared error cost function trains weights that span the same subspace as the one spanned by the principal component loading vectors, but that they are not identical to the loading vectors. In this paper, we show how to recover the loading vectors from the autoencoder weights.
\end{abstract}


%
\IEEEpeerreviewmaketitle

\section{Background}
\IEEEPARstart{P}{rincipal Component Analysis} (PCA) is a linear transformation that transforms a set of observations to a new coordinate system in which the values of the first coordinate have the largest possible variance, and the values of each succeeding coordinate have the largest possible variance under the constraint that they are uncorrelated with the preceding coordinates. They are often found by either computing the eigendecomposition of the covariance matrix or by computing the singular value decomposition of the observations.

By keeping only the first few principal components, PCA can be used for dimensionality reduction. The decorrelation of the coordinates is also a useful property, and PCA is sometimes used as a preprocessing step for whitening a dataset before using it as an input into an optimization problem such as a neural network classifier.

One of the properties of PCA is that out of all possible linear transformations, the reconstructions of the observations from the leading principal components have the least total squared error. Autoencoders are neural networks that aim to minimize the error of reconstructions of observations. It is well known that an autoencoder with a single fully-connected hidden layer, a linear activation function and a squared error cost function is closely related to PCA - its weights span the principal subspace, which is the subspace spanned by the first loading vectors \cite{AutoencoderSVD}, \cite{WithoutLocalMinima}. However, they are not equal to the loading vectors.

This paper proposes a simple method for recovering the loading vectors from the weights of a linear autoencoder. This allows the usage of autoencoders for computing PCA, unlike existing methods which merely find the principal subspace. After recovering the loading vectors, the solution has the following properties: (i) it is unique; (ii) in the transformed data, different coordinates are uncorrelated; (iii) the coordinates are sorted in descending order of variance; and (iv) the solutions for reduction to different dimensions are nested: when reducing the data from dimension $n$ to dimension $m_1$, the first $m_2$ vectors ($m_2 < m_1$) are the same as the solution for reduction from dimension $n$ to $m_2$. These properties do not hold for a general basis for the principal subspace, which is what the autoencoder weights converge to.

Even though any method for computing PCA using autoencoders is highly inefficient for small datasets, for large datasets such a method is desirable because of its various advantages compared to the standard methods. Autoencoders can be trained by a variety of stochastic optimization methods that have been developed for training deep neural networks. These optimizers can handle high-dimensional training data such as images, and a large number of them. They are suitable for the online learning scenario in which new data arrives over time, and they do not require subtracting from each example the element-wise mean of the entire training set. While several online PCA methods have been proposed for meeting these demands \cite{Online1}, \cite{Online2}, \cite{Online3}, our method is the first to simply recover the loading vectors from the weights of an autoencoder.

\section{Introduction}
\subsection{Principal Component Analysis}
Let $\left\{ \textbf{y}_i \right\}_{i=1}^N$ be a set of $N$ observation vectors, each of dimension $n$. We assume that $n \leq N$, which is the more common scenario in machine learning, but a similar analysis can be done for the case of $n > N$. Let $\text{Y} \in \mathbb{R}^{n \times N}$ be a matrix whose columns are $\left\{ \textbf{y}_i \right\}_{i=1}^N$,
\[
\text{Y} = \left[\begin{array}{ccc}
\mid &  & \mid\\
\textbf{y}_{1} & \cdots & \textbf{y}_{N}\\
\mid &  & \mid
\end{array}\right].
\]
The element-wise average of the $N$ observations is an $n$ dimensional signal which may be written as:
\[
\bar{\textbf{y}} = \frac{1}{N}\overset{N}{\underset{i=1}{\sum}} \textbf{y}_i = \frac{1}{N} \text{Y} \mathbb{1}_N,
\]
where $\mathbb{1}_N \in \mathbb{R}^{N \times 1}$ is a column vector of all-ones. Let $\text{Y}_0$ be a matrix whose columns are the centered observations (we center each observation $\textbf{y}_i$ by subtracting $\bar{\textbf{y}}$ from it): 
\[
\text{Y}_0 = \text{Y} - \bar{\textbf{y}} \mathbb{1}_N^T.
\]

A linear transformation of a finite dimensional vector may be expressed as a matrix multiplication:
\[
\textbf{x}_i = \text{W}^T\textbf{y}_i,
\]
where $\textbf{y}_i \in \mathbb{R}^n$, $\textbf{x}_i \in \mathbb{R}^m$, and $\text{W} \in \mathbb{R}^{n \times m}$.
Each element $j$ in the vector $\textbf{x}_i$ is an inner product between $\textbf{y}_i$ and the $j$-th column of $\text{W}$, which we denote by $\textbf{w}_j$.

Let $\text{X} \in \mathbb{R}^{m \times N}$ be a matrix whose columns are the set of $N$ vectors of transformed observations, let $\bar{\textbf{x}} = \frac{1}{N}\overset{N}{\underset{i=1}{\sum}} \textbf{x}_i = \frac{1}{N} \text{X} \mathbb{1}_N$ be the element-wise average, and $\text{X}_0 = \text{X} - \bar{\textbf{x}} \mathbb{1}_N^T$ the centered matrix. Clearly, $\text{X} = \text{W}^T \text{Y}$ and $\text{X}_0 = \text{W}^T \text{Y}_0$. 

When the matrix $\text{W}^T$ represents the transformation that applies principal component analysis, we denote $\text{W}=\text{P}$, and the columns of $\text{P}$, denoted $\left\{ \textbf{p}_j \right\}_{j=1}^n$, are referred to as \emph{loading vectors}. The transformed vectors $\left\{ \textbf{x}_i \right\}_{i=1}^N$ are referred to as \emph{principal components} or \emph{scores}. The first loading vector is defined as the unit vector with which the inner products of the observations have the greatest variance: 
\begin{equation}
\textbf{p}_1 = \underset{\textbf{w}_1}{\max}{~\textbf{w}_1^T \text{Y}_0 \text{Y}_0^T \textbf{w}_1 }~s.t.~\textbf{w}_1^T \textbf{w}_1 = 1.
\label{eq:maxvar}
\end{equation}
The solution to (\ref{eq:maxvar}) is known to be the eigenvector of the sample covariance matrix $\text{Y}_0 \text{Y}_0^T$ corresponding to its largest eigenvalue. We normalize the eigenvector and disregard its sign. Next, $\textbf{p}_2$ is the unit vector which has the largest variance of inner products between it and the observations after removing the orthogonal projections of the observations onto $\textbf{p}_1$. It may be found by solving:
\begin{multline}
\textbf{p}_2 = \underset{\textbf{w}_2}{\max}{~\textbf{w}_2^T \left( \text{Y}_0 - \textbf{p}_1 \textbf{p}_1^T \text{Y}_0 \right) \left( \text{Y}_0 - \textbf{p}_1 \textbf{p}_1^T \text{Y}_0 \right)^T \textbf{w}_2 }\\~s.t.~ \textbf{w}_2^T \textbf{w}_2 = 1.
\label{eq:p2}
\end{multline}
The solution to (\ref{eq:p2}) is known to be the eigenvector corresponding to the largest eigenvalue under the constraint that it is not collinear with $\textbf{p}_1$. Similarly, the remaining loading vectors are equal to the remaining eigenvectors of $\text{Y}_0 \text{Y}_0^T$ corresponding to descending eigenvalues. The eigenvalues of $\text{Y}_0 \text{Y}_0^T$, which is a positive semi-definite matrix, are non-negative. They are not necessarily distinct, but since it is a symmetric matrix it has $n$ eigenvectors that are all orthogonal, and it is always diagonalizable. Thus, the matrix $\text{P}$ may be computed by diagonalizing the covariance matrix:
\[
\text{Y}_0 \text{Y}_0^T = \text{P} \Lambda \text{P}^{-1} = \text{P} \Lambda \text{P}^T,
\]
where $\Lambda = \text{X}_0 \text{X}_0^T$ is a diagonal matrix whose diagonal elements $\left\{ \lambda_i \right\}_{i=1}^n$ are sorted in descending order, the columns of $\text{P}$ are orthonormal, i.e. $\text{P}^{-1}=\text{P}^T$. The transformation back to the observations is $\text{Y} = \text{P} \text{X}$. The fact that the covariance matrix of $\text{X}$ is diagonal means that PCA is a decorrelation transformation. By dividing each coordinate by the square root of its corresponding eigenvalue, PCA can be used as a whitening transformation, which is sometimes used as a preprocessing step in order to cause optimization problems to converge more easily.

\subsection{Dimensionality reduction}
PCA is often used as a method for dimensionality reduction, the process of reducing the number of variables in a model in order to avoid the curse of dimensionality. This is done by simply keeping the first $m$ principal components ($m<n$), i.e., applying the truncated transformation
\[
\text{X}_m = \text{P}_m^T \text{Y},
\]
where each column of $\text{X}_m \in \mathbb{R}^{m \times N}$ is a vector whose elements are the first $m$ principal components, and $\text{P}_m$ is a matrix whose columns are the first $m$ loading vectors,
\[
\text{P}_m = \left[\begin{array}{ccc}
\mid &  & \mid\\
\textbf{p}_{1} & \cdots & \textbf{p}_{m}\\
\mid &  & \mid
\end{array}\right] \in \mathbb{R}^{n \times m}.
\]
Intuitively, by keeping only $m$ principal components, we are losing information, and we minimize this loss of information by maximizing their variances. Many iterative algorithms (e.g., QR algorithm, Jacobi algorithm, and the power method) can efficiently find the $m$ largest eigenvalues of $\text{Y}_0 \text{Y}_0^T$ and their corresponding eigenvectors without having to fully diagonalize the matrix. Yet, for high dimensional data, computing $\text{Y}_0 \text{Y}_0^T$ may be prohibitive.

\subsection{Singular-value decomposition}
Any matrix $\text{Y}_0 \in \mathbb{R}^{n \times N}$ may be factorized as $\text{Y}_0 = \text{U} \Sigma \text{V}^T$, where $\text{U} \in \mathbb{R}^{n \times n}$ and $\text{V} \in \mathbb{R}^{N \times N}$ are both orthogonal matrices and $\Sigma \in \mathbb{R}^{n \times N}$ is a matrix whose elements are non-negative real numbers on the diagonal and zero elsewhere. The diagonal elements $\left\{ \sigma_i \right\}_{i=1}^n$, referred to as singular values, are sorted in descending order, and the columns of $\text{U}$ and $\text{V}$ are referred to, respectively, as left and right singular vectors. Assuming $n \leq N$ and $\text{Y}_0$ is full-rank, the columns of $\text{U}$ are an orthonormal basis for $\mathbb{R}^n$. If $n > N$ (the number of observation is smaller than the dimension of each observation), then the first $n$ columns of $\text{U}$ are an orthonormal basis for the column space of $\text{Y}_0$, and the remaining $N-n$ columns are an orthonormal basis for its nullspace.

The covariance matrix may be written as:
\[
\text{Y}_0 \text{Y}_0^T = \text{U} \Sigma \text{V}^T \text{V} \Sigma^T  \text{U}^T =  \text{U} \Sigma \Sigma^T \text{U}^{-1},
\]
where $\Sigma \Sigma^T$ is a diagonal matrix. Thus, singular-value decomposition of $\text{Y}_0$ is equivalent to eigendecomposition of $\text{Y}_0 \text{Y}_0^T$. The singular values of $\text{Y}_0$ are the square roots of the eigenvalues of $\text{Y}_0 \text{Y}_0^T$, and the left singular vectors of $\text{Y}_0$ are the eigenvectors of $\text{Y}_0 \text{Y}_0^T$. Eigendecomposition is unique up to the scale of the eigenvectors, which we normalize, and to permutations of the eigenvectors and their corresponding eigenvalues, which we sort in descending order. Therefore, the left singular vectors of $\text{Y}_0$ must be equal to the loading vectors of $\text{Y}$ (up to their sign, which we disregard).

Iterative algorithms (e.g., QR algorithm, one-sided Jacobi algorithm) can efficiently find the $m$ largest singular values and their corresponding left singular vectors without having to fully decompose the matrix.

\subsection{PCA and large sets of high-dimensional data}
When the number of observations $N$ is small enough to fit in memory, SVD is often the preferred method for computing the loading vectors, as it avoids computing the covariance matrix $\text{Y}_0 \text{Y}_0^T$, which is desirable especially when $n$ is large. 

When the number of observations $N$ is large but the dimension $n$ is sufficiently small, $\text{Y}_0 \text{Y}_0^T = \overset{N}{\underset{i=1}{\sum}} \left(\textbf{y}_i-\bar{\textbf{y}}\right) \left(\textbf{y}_i-\bar{\textbf{y}}\right)^T$ may be computed sequentially, with a memory requirement of $O \left( n^2 \right)$ instead of $O \left( nN \right)$. Thus, there is no need to load the entire dataset into memory.

However, when the number of observations is large and each observation is high-dimensional (i.e., both $N$ and $n$ are large), this too may be infeasible. Online methods \cite{Online1}, \cite{Online2}, \cite{Online3} iterate through the dataset one example at a time or in minibatches. When applied to images, the images are more often divided into small patches, and PCA is applied to patches rather than to the entire images; this is known as local PCA \cite{LocalPCA}.

\subsection{Minimum total squared reconstruction error}
Interestingly, $\text{P}_m$ is also a solution to:
\begin{equation}
\underset{\text{W} \in \mathbb{R}^{n \times m}}{\text{min}}{~\left\Vert \text{Y}_0-\text{W} \text{W}^T \text{Y}_0 \right\Vert_F^2}~s.t.~ \text{W}^T \text{W} = \text{I}_{m \times m},
\label{eq:MMSE}
\end{equation}
where $F$ denotes the Frobenius norm. According to this formulation, the $m$ leading loading vectors are an orthonormal basis which spans the $m$ dimensional subspace onto which the projections of the centered observations have the minimum squared difference from the original centered observations. In other words, $\text{P}_m$ compresses each centered vector of length $n$ into a vector of length $m$ (where $m \leq n$) in such a way that minimizes the sum of total squared reconstruction errors.

It is well known \cite{EckardtYoung} and easily verified that $\text{P}_m$ indeed solves (\ref{eq:MMSE}). Yet, the minimizer of (\ref{eq:MMSE}) is not unique: $\text{W} = \text{P}_m \text{Q}$ is also a solution, where $\text{Q} \in \mathbb{R}^{m \times m}$ is any orthogonal matrix, $\text{Q}^T = \text{Q}^{-1}$. Multiplying $\text{P}_m$ from the right by $\text{Q}$ transforms the first $m$ loading vectors into a different orthonormal basis for the same subspace.

\subsection{Autoencoders}
A neural network that is trained to learn the identity function is called an autoencoder. Its output layer has the same number of nodes as the input layer, and the cost function is some measure of the reconstruction error. Autoencoders are unsupervised learning models, and they are often used for the purpose of dimensionality reduction \cite{Dimreduction}. A simple autoencoder that implements dimensionality reduction is a feed-forward autoencoder with at least one layer that has a smaller number of nodes, which functions as a bottleneck. After training the neural network using backpropagation, it is separated into two parts: the layers up to the bottleneck are used as an encoder, and the remaining layers are used as a decoder.

In the simplest case, there is only one hidden layer (the bottleneck), and the layers in the network are fully connected. A vector $\textbf{y}_i \in \mathbb{R}^{n \times 1}$ passes through the hidden layer, which outputs $\textbf{x}_i \in \mathbb{R}^{m \times 1}$ according to the mapping $\textbf{x}_i = a\left( \text{W}_1 \textbf{y}_i + \textbf{b}_1 \right)$, where $\text{W}_1 \in \mathbb{R}^{m \times n}$ is referred to as the weight matrix of the first layer, $\textbf{b}_1 \in \mathbb{R}^{m \times 1}$ is referred to as the bias vector of the first layer, and $m < n$. The function $a$, referred to as the activation function, operates element-wise and is typically a non-linear function such as the rectified linear unit (ReLU). The second layer maps $\textbf{x}_i$ to $\hat{\textbf{y}}_i \in \mathbb{R}^{n \times 1}$ according to $\hat{\textbf{y}}_i = a\left( \text{W}_2 \textbf{x}_i + \textbf{b}_2 \right) = a\left( \text{W}_2 a\left( \text{W}_1 \textbf{y}_i + \textbf{b}_1 \right)+ \textbf{b}_2 \right)$, where $\text{W}_2 \in \mathbb{R}^{n \times m}$ and $\textbf{b}_2 \in \mathbb{R}^{n \times 1}$ are the weight matrix and bias vector of the second layer. The parameters $\text{W}_1, \textbf{b}_1, \text{W}_2, \textbf{b}_2$ are found by minimizing some cost function measuring the difference between the output $\hat{\textbf{y}}_i$ and the input $\textbf{y}_i$. Using backpropagation with an optimizer such as stochastic gradient descent, each data sample from $\left\{ \textbf{y}_i \right\}_{i=1}^N$ is fed through the network to compute $\textbf{x}_i$ and $\hat{\textbf{y}}_i$, which are then used to compute the gradients and to update the parameters.

\subsection{Linear autoencoders}
In the case that no non-linear activation function is used, $\textbf{x}_i = \text{W}_1 \textbf{y}_i + \textbf{b}_1$ and $\hat{\textbf{y}}_i = \text{W}_2 \textbf{x}_i + \textbf{b}_2$. If the cost function is the total squared difference between output and input, then training the autoencoder on the input data matrix $\text{Y}$ solves:
\begin{equation}
\underset{\text{W}_1, \textbf{b}_1, \text{W}_2, \textbf{b}_2 }{\text{min}}{~
\left\Vert \text{Y} - \left(
\text{W}_2 \left(\text{W}_1 \text{Y} + \textbf{b}_1 \mathbb{1}_N^T \right) + \textbf{b}_2 \mathbb{1}_N^T
\right) \right\Vert_F^2 }.
\label{eq:NNMSE}
\end{equation}
In \cite{AutoencoderSVD}, it is shown that if we set the partial derivative with respect to $\textbf{b}_2$ to zero and insert the solution into (\ref{eq:NNMSE}), then the problem becomes:
\[
\underset{\text{W}_1, \text{W}_2 }{\text{min}}{~\left\Vert \text{Y}_0 - \text{W}_2 \text{W}_1 \text{Y}_0 \right\Vert_F^2}
\]
Thus, for any $\textbf{b}_1$, the optimal $\textbf{b}_2$ is such that the problem becomes independent of $\textbf{b}_1$ and of $\bar{\textbf{y}}$. Therefore, we may focus only on the weights $\text{W}_1$, $\text{W}_2$. In \cite{WithoutLocalMinima} it is shown that when setting the gradients to zero, $\text{W}_1$ is the left Moore-Penrose pseudoinverse of $\text{W}_2$ (and $\text{W}_2$ is the right pseudoinverse of $\text{W}_1$):
\[
\text{W}_1 = \text{W}_2^{\dagger} = \left( \text{W}_2^T \text{W}_2 \right)^{-1} \text{W}_2^T
\]
Thus, the minimization remains with respect to a single matrix:
\begin{equation}
\underset{\text{W}_2 \in \mathbb{R}^{n \times m}}{\text{min}}{~\left\Vert \text{Y}_0 - \text{W}_2 \text{W}_2^{\dagger} \text{Y}_0 \right\Vert_F^2}
\label{eq:NNsubspace}
\end{equation}
The matrix $\text{W}_2 \text{W}_2^{\dagger} = \text{W}_2 \left( \text{W}_2^T \text{W}_2 \right)^{-1} \text{W}_2^T$ is the orthogonal projection operator onto the column space of $\text{W}_2$ when its columns are not necessarily orthonormal. This problem is very similar to (\ref{eq:MMSE}), but without the orthonormality constraint.

In \cite{AutoencoderSVD} and \cite{WithoutLocalMinima} it is shown that $\text{W}_2$ is a minimizer of (\ref{eq:NNsubspace}) if and only if its column space is spanned by the first $m$ loading vectors of $\text{Y}$. This can also be shown by applying the QR decomposition to $\text{W}_2$, which transforms the problem (\ref{eq:NNsubspace}) into the one in (\ref{eq:MMSE}). As a result, it is possible to solve (\ref{eq:MMSE}) by first solving the unconstrained problem (\ref{eq:NNsubspace}), and then orthonormalizing the columns of the solution, e.g. using the Gram-Schmidt process. However, this does not recover the loading vectors $\text{P}_m$, but rather $\text{P}_m \text{Q}$ for some unknown orthogonal matrix $\text{Q}$.

The linear autoencoder is said to apply PCA to the input data in the sense that its output is a projection of the data onto the low dimensional principal subspace. However, unlike actual PCA, the coordinates of the output of the bottleneck are correlated and are not sorted in descending order of variance. In addition, the solutions for reduction to different dimensions are not nested: when reducing the data from dimension $n$ to dimension $m_1$, the first $m_2$ vectors ($m_2 < m_1$) are not an optimal solution to reduction from dimension $n$ to $m_2$, which therefore requires training an entirely new autoencoder.

Several methods have been proposed for neural networks that compute the exact loading vectors \cite{NNPCA1}, \cite{NNPCA2}, \cite{NNPCA3}, \cite{NNPCA4}, \cite{NNPCA5}. However, they require specific algorithms for iteratively updating the weights, and as such are similar to online PCA methods. No method has so far been proposed for recovering the loading vectors from a simple linear autoencoder that is independent of the optimization method used for training it.

\section{Method}

\emph{Hypothesis}: The first $m$ loading vectors of $\text{Y}$ are the first $m$ left singular vectors of the matrix $\text{W}_2$ which minimizes (\ref{eq:NNsubspace}).

\emph{Note: an earlier version of this work attempted to prove this statement, but was found to be erroneous.}

In other words, instead of computing the first $m$ left singular vectors of $\text{Y}_0 \in \mathbb{R}^{n \times N}$, we may train a linear autoencoder on the (non-centered) dataset $\text{Y}$ and then compute the first $m$ left singular vectors of $\text{W}_2 \in \mathbb{R}^{n \times m}$, where typically $m << N$. The loading vectors may also be recovered from the weights of the hidden layer, $\text{W}_1$. If $\text{W}_2 = \text{U} \Sigma \text{V}^T$, then $\text{W}_1 = \text{W}_2^{\dagger} = \text{V} \Sigma^{\dagger} \text{U}^T$, and $\text{W}_1^T = \text{U} \left( \Sigma^{\dagger} \right)^T \text{V}^T$. Thus, the first $m$ left singular vectors of $\text{W}_1^T \in \mathbb{R}^{n \times m}$ are also equal to the first $m$ loading vectors of $\text{Y}$.

\section{Experiments}
\begin{figure}[!b]
\centering
\includegraphics[width=3.2in]{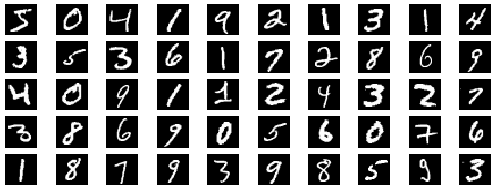}
\caption{Examples of images from the MNIST dataset.}
\label{fig0}
\end{figure}
\begin{figure*}
    \centering
  \subfloat[$\text{P}_m$]{%
       \includegraphics[width=0.3\linewidth]{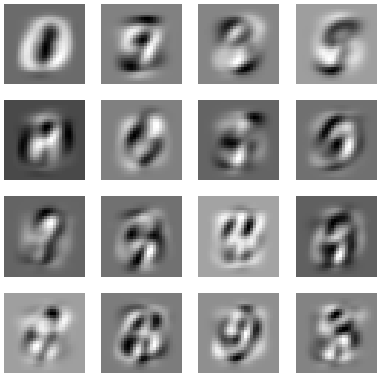}}
    \label{1a}\hfill
  \subfloat[$\text{W}_2$]{%
        \includegraphics[width=0.3\linewidth]{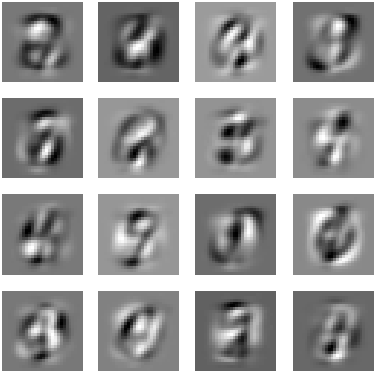}}
    \label{1b}\hfill
  \subfloat[$\text{U}_m$]{%
        \includegraphics[width=0.3\linewidth]{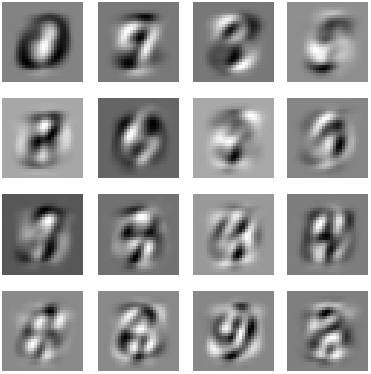}}
    \label{1c}
  \caption{(a) The first 16 loading vectors of MNIST, computed by applying SVD to the entire dataset, (b) the weights of a linear autoencoder trained on the dataset, (c) the left singular vectors of the autoencoder weights. Notice how the left singular vectors in (c) are very close to the loading vectors in (a) up to their sign (inversion of the gray levels in some of the images).}
  \label{fig1} 
\end{figure*}

\begin{figure*} 
    \centering
  \subfloat[$\text{P}_m^T \text{Y}_0 \text{Y}_0^T \text{P}_m$]{%
       \includegraphics[width=0.28\linewidth]{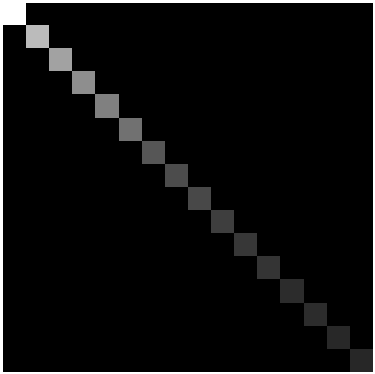}}
    \label{1a}\hfill
  \subfloat[$\text{W}_2^T \text{Y}_0 \text{Y}_0^T \text{W}_2$]{%
        \includegraphics[width=0.28\linewidth]{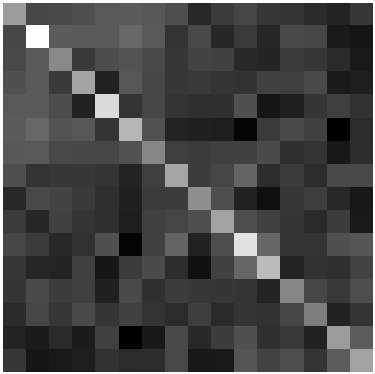}}
    \label{1b}\hfill
  \subfloat[$\text{U}_m^T \text{Y}_0 \text{Y}_0^T \text{U}_m$]{%
        \includegraphics[width=0.28\linewidth]{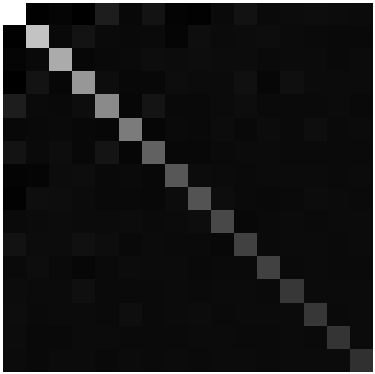}}
    \label{1c}
  \caption{The covariance matrix of the data in the transformed coordinates, according to (a) the loading vectors computed by applying SVD to the entire dataset, (b) the weights of the linear autoencoder, and (c) the left singular vectors of the autoencoder weights.}
  \label{fig2}
\end{figure*}

An important advantage of applying PCA using a linear autoencoder is that it is very simple to implement using popular machine learning frameworks. We release a sample implementation of a linear autoencoder and the recovery of the loading vectors from its weights which uses the Keras library. It can be found at \emph{https://github.com/plaut/linear-ae-pca}.

As in any neural network, we initialize the values of the parameters, including $\text{W}_2$, to random numbers. Notice that the optimization is convex over $\text{W}_2$ for a fixed $\text{W}_1$ and it is convex over $\text{W}_1$ for a fixed $\text{W}_2$, but it is not jointly convex and has many saddle points which may be far from optimal. The optimization may benifit from a more sophisticated update than basic stochastic gradient descent, and we chose to use the Adam optimizer. Weight decay regularization, which penalizes unreasonable factorizations, was also found to be beneficial.

\subsection{MNIST}
We trained a linear autoencoder on the MNIST training dataset \cite{MNISTdata}, which contains 60,000 grayscale images of handwritten digits, each of size $28 \times 28$. Fig. \ref{fig0} shows a few examples of images from the dataset. We set the dimensions of the network for reduction from a dimension of $28 \times 28=784$ to a dimension of $16$. Then, we applied our method for recovering the loading vectors from the weights of the autoencoder.

For comparison, Fig. \ref{fig1}(a) shows the first 16 loading vectors as computed using the standard method of applying SVD to the entire dataset after centering it. Fig. \ref{fig1}(b) shows the columns of $\text{W}_2$, and Fig. \ref{fig1}(c) shows the columns of $\text{U}_m$ (the first $m$ left singular vectors of $\text{W}_2$). It is evident that although the columns of $\text{W}_2$ are entirely different from the loading vectors, their left singular vectors are approximately equal to them up to sign. Seven of the vectors have opposite signs, which is reflected in their inverted gray levels.

After computing the loading vectors, we applied three different transformations to the centered: $\text{P}_m^T \text{Y}_0$, $\text{W}_2^T \text{Y}_0$, and $\text{U}_m^T \text{Y}_0$. Fig. \ref{fig2} shows the covariance matrix in the transformed coordinates for the three transformations. As expected, $\text{U}_m^T$ transformed the data to coordinates in which the covariance is a diagonal matrix with descending diagonal elements (similarly to $\text{P}_m^T$), while $\text{W}_2^T$ did not.

In order to further reduce the dimensionality to $m_2 < 16$, all we need to do is keep the first $m_2$ rows of $\text{U}_m^T \text{Y}_0$.

\subsection{CUB-200-2011}
\begin{figure}[!t]
\centering
\includegraphics[width=3.4in]{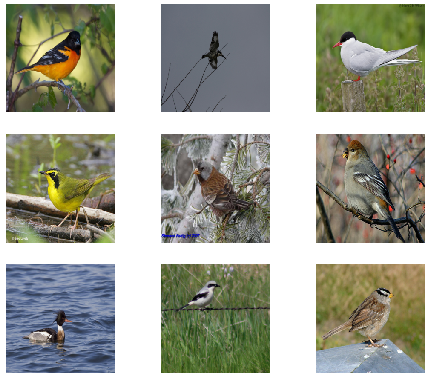}
\caption{Examples of images from the CUB-200-2011 dataset.}
\label{fig2.5}
\end{figure}

\begin{figure}[!t]
\centering
\includegraphics[width=3.6in]{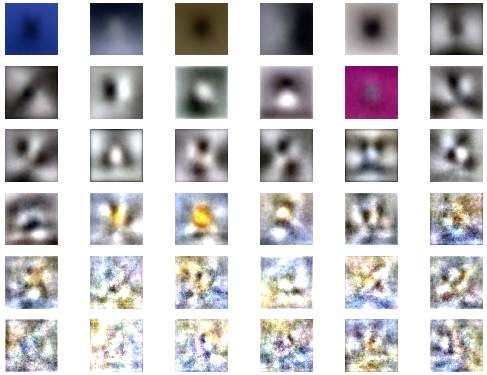}
\caption{The loading vectors of the CUB-200-2011 dataset, as recovered from the weight matrix $\text{W}_2$.}
\label{fig3}
\end{figure}

Computing PCA using a linear autoencoder only becomes advantageous when handling a large set of large images. We applied the same technique to the CUB-200-2011 dataset \cite{WahCUB}, which contains 11,788 color images of birds. The images were resized to $256 \times 256$. Fig. \ref{fig2.5} shows a few examples of images from the dataset. The autoencoder was set for dimensionality reduction from a dimension of $256 \times 256 \times 3 = 196,608$ to a dimension of 36. Then, the first 36 loading vectors of the dataset were recovered by applying SVD to the weight matrix $\text{W}_2$ and taking the first 36 left singular vectors of it, they are shown in Fig. \ref{fig3}.

\begin{figure}[!t]
\centering
\includegraphics[width=2.5in]{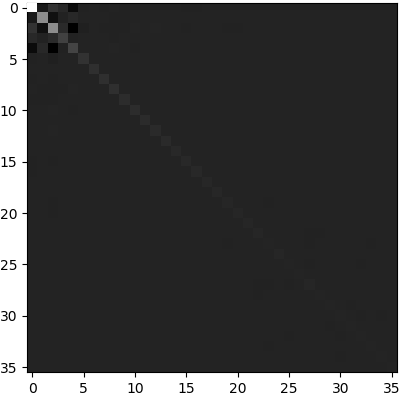}
\caption{The covariance matrix of the CUB-200-2011 dataset in the transformed coordinates. As expected, the covariance is approximately diagonal with descending elements on the diagonal.}
\label{fig4}
\end{figure}

In order to verify that the resulting transformation applies PCA, we centered the dataset and calculated the covariance in the transformed coordinates, shown in Fig. \ref{fig4}. As expected, it is approximately a diagonal matrix with descending elements on the diagonal.

This dataset was too large to fit in memory, so we did not compare the results to applying SVD to the entire dataset. One might suspect that the last loading vectors in Fig. \ref{fig3} are evidence of underfitting due to their high spatial frequencies. In order to rule this out, we computed the loading vectors of a subset of 1,000 examples from the dataset by applying SVD to all the 1,000 examples after centering them, and they exhibited the same appearance.


\section{Conclusion}
It is in fact possible to use a linear autoencoder not only to project data onto the principal subspace, but to actually perform principal component analysis. Recovering the loading vectors amounts to simply applying SVD to the weight matrix of one of the two layers. The solution is independent of the optimization algorithm used to train the neural network. The advantages of implementing PCA this way are that is able to:
\begin{itemize}
\item Process high-dimensional data.
\item Process datasets with large numbers of observations.
\item Avoid the need to center the data to element-wise zero mean.
\item Process new data online as it arrives.
\item Be implemented easily using very few lines of code, using the latest algorithms, software frameworks and hardware that are optimized for training neural networks.
\end{itemize}


%






\ifCLASSOPTIONcaptionsoff
  \newpage
\fi



\bibliographystyle{IEEEtran}

\begin{thebibliography}{10}
\providecommand{\url}[1]{#1}
\csname url@samestyle\endcsname
\providecommand{\newblock}{\relax}
\providecommand{\bibinfo}[2]{#2}
\providecommand{\BIBentrySTDinterwordspacing}{\spaceskip=0pt\relax}
\providecommand{\BIBentryALTinterwordstretchfactor}{4}
\providecommand{\BIBentryALTinterwordspacing}{\spaceskip=\fontdimen2\font plus
\BIBentryALTinterwordstretchfactor\fontdimen3\font minus
  \fontdimen4\font\relax}
\providecommand{\BIBforeignlanguage}[2]{{%
\expandafter\ifx\csname l@#1\endcsname\relax
\typeout{** WARNING: IEEEtran.bst: No hyphenation pattern has been}%
\typeout{** loaded for the language `#1'. Using the pattern for}%
\typeout{** the default language instead.}%
\else
\language=\csname l@#1\endcsname
\fi
#2}}
\providecommand{\BIBdecl}{\relax}
\BIBdecl

\bibitem{AutoencoderSVD}
H.~Bourlard and Y.~Kamp, ``Auto-association by multilayer perceptrons and
  singular value decomposition,'' \emph{Biological Cybernetics}, vol. 59(4-5),
  pp. 291--294, 1988.

\bibitem{WithoutLocalMinima}
P.~Baldi and K.~Hornik, ``Neural networks and principal component analysis:
  Learning from examples without local minima,'' \emph{Neural Networks}, vol.
  2(1), pp. 53--58, 1989.

\bibitem{Online1}
M.~Warmuth and D.~Kuzmin, ``Randomized online {PCA} algorithms with regret
  bounds that are logarithmic in the dimension,'' \emph{Journal of Machine
  Learning Research}, vol.~9, pp. 2287--2320, 2008.

\bibitem{Online2}
J.~Fend, H.~Xu, S.~Mannor, and S.~Yan, ``Online {PCA} for contaminated data,''
  \emph{Advances in Neural Information Processing Systems}, pp. 764--772, 2013.

\bibitem{Online3}
J.~Fend, H.~Xu, and S.~Yan, ``Online robust {PCA} via stochastic
  optimization,'' \emph{Advances in Neural Information Processing Systems}, pp.
  404--412, 2013.

\bibitem{LocalPCA}
N.~Kambhatla and T.~Leen, ``Dimension reduction by local principal component
  analysis,'' \emph{Neural Computation}, vol. 9(7), pp. 1493--1516, 1997.

\bibitem{EckardtYoung}
C.~Eckardt and G.~Young, ``The approximation of one matrix by another of lower
  rank,'' \emph{Psychometrika}, vol.~1, pp. 211--218, 1936.

\bibitem{Dimreduction}
G.~Hinton and R.~Salakhutdinov, ``Reducing the dimensionality of data with
  neural networks,'' \emph{Science}, vol. 313(5786), pp. 504--507, 2006.

\bibitem{NNPCA1}
E.~Oja, ``Neural networks, principal components, and subspaces,''
  \emph{International Journal of Neural Systems}, vol. 1(01), pp. 61--68, 1989.

\bibitem{NNPCA2}
J.~Rubner and P.~Tavan, ``A self-organizing network for principal-component
  analysis,'' \emph{EPL (Europhysics Letters)}, vol. 10(7), p. 693, 1989.

\bibitem{NNPCA3}
S.~Kung and K.~Diamantaras, ``A neural network learning algorithm for adaptive
  principal component extraction ({APEX}),'' \emph{ICASSP}, pp. 861--864, 1990.

\bibitem{NNPCA4}
E.~Oja, ``Principal components, minor components, and linear neural networks,''
  \emph{Neural Networks}, vol. 5(6), pp. 927--935, 1992.

\bibitem{NNPCA5}
L.~Xu, ``Least mean square error reconstruction principle for self-organizing
  neural-nets,'' \emph{Neural Networks}, vol. 6(5), pp. 627--648, 1993.

\bibitem{Freedman}
D.~A. Freedman, ``Statistical models: Theory and practice.'' \emph{Cambridge
  University Press}, 2009.

\bibitem{Antoulas}
A.~Antoulas, ``Approximation of large-scale dynamical systems,'' \emph{SIAM},
  vol.~6, pp. 37--38, 2005.

\bibitem{MNISTdata}
Y.~LeCun, C.~Cortes, and C.~Burges, ``{MNIST} handwritten digit database,''
  2010.

\bibitem{WahCUB}
C.~Wah, S.~Branson, P.~Welinder, P.~Perona, and S.~Belongie, ``The
  {Caltech}-{UCSD} birds-200-2011 dataset,'' no. CNS-TR-2011-001, 2011.

\end{thebibliography}
\end{document}